\documentclass[10pt,twocolumn,letterpaper]{article}

\usepackage{iccv}
\usepackage{times}
\usepackage{epsfig}
\usepackage{graphicx}
\usepackage{amsmath}
\usepackage{amssymb}
\usepackage{amsfonts}
\usepackage{subcaption} 
\usepackage{booktabs}
\usepackage{caption}
\newcommand{\norm}[1]{\| #1 \|}
\usepackage{multirow}
\usepackage{soul}
\usepackage{enumitem}

\usepackage[pagebackref=true,breaklinks=true,letterpaper=true,colorlinks,bookmarks=false]{hyperref}

\iccvfinalcopy % *** Uncomment this line for the final submission

\def\iccvPaperID{9262} % *** Enter the ICCV Paper ID here

\ificcvfinal\fi

\begin{document}
	
	%%%%%%%%% TITLE
	\title{GPRAR: \underline{G}raph Convolutional Network based \underline{P}ose \underline{R}econstruction and \underline{A}ction \underline{R}ecognition for Human Trajectory Prediction}
	
	\author{Manh Huynh and Gita Alaghband\\
		University of Colorado Denver\\
		{\tt\small \{manh.huynh, gita.alaghband\}@ucdenver.edu}
		% \and
		%Gita Alaghband\\
		%%University of Colorado Denver\\
	}
	
	\maketitle
	\ificcvfinal\thispagestyle{empty}\fi
	
	%%%%%%%%% ABSTRACT
	\begin{abstract}
		Prediction with high accuracy is essential for various applications such as autonomous driving. Existing prediction models are easily prone to errors in real-world settings where observations (e.g. human poses and locations) are often noisy. To address this problem, we introduce GPRAR, a graph convolutional network based pose reconstruction and action recognition for human trajectory prediction. The key idea of GPRAR is to generate robust features: human poses and actions, under noisy scenarios. To this end, we design GPRAR using two novel sub-networks: PRAR (Pose Reconstruction and Action Recognition) and FA (Feature Aggregator). PRAR aims to simultaneously reconstruct human poses and action features from the coherent and structural properties of human skeletons. It is a network of an encoder and two decoders, each of which comprises multiple layers of spatiotemporal graph convolutional networks. Moreover, we propose a Feature Aggregator (FA) to channel-wise aggregate the learned features: human poses, actions, locations, and camera motion using encoder-decoder based temporal convolutional neural networks to predict future locations. Extensive experiments on the commonly used datasets: JAAD~\cite{kotseruba2016joint} and TITAN~\cite{malla2020titan} show accuracy improvements of GPRAR over state-of-the-art models. Specifically, GPRAR improves the prediction accuracy up to $22\%$ and $50\%$ under noisy observations on JAAD and TITAN datasets, respectively.
	\end{abstract}
	
	%%%%%%%%% BODY TEXT
	\section{Introduction}
	Accurate prediction of human trajectory, i.e., forecasting pedestrians’ future locations given their past (observed) frames in dynamic scenes, is critical for various applications such as autonomous driving~\cite{levinson2011towards}, robotic navigation systems~\cite{luong2021incremental}, and pedestrian tracking~\cite{manh2018spatiotemporal}. For the most part, challenges associated with predicting future trajectories are due to the presence of a multitude of features that may influence human future paths such as camera motion (egomotion), human shapes (pose), past locations, and human actions. More importantly, these features are often noisy due to environmental and scene impediments, occlusions for example. This problem has significantly degraded the performance of feature extractors, which in turn degrades the accuracy of the existing prediction models.  
	\begin{figure}[t]
		\centering
		%\fbox{\rule{0pt}{2in} \rule{.9\linewidth}{0pt}}
		\includegraphics[width=0.5\textwidth]{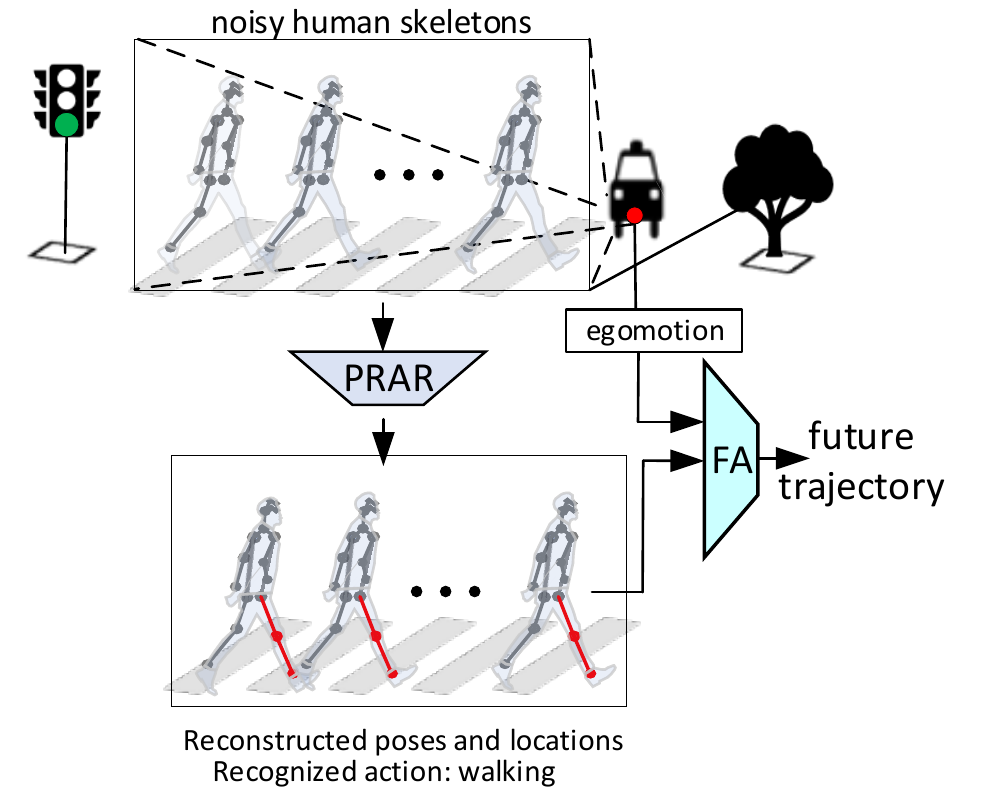}
		
		\caption{GPRAR  addresses the problem of future trajectory prediction given noisy pose observations by two novel sub-networks: (1) a human Pose Reconstruction and Action Recognition network (PRAR) to simultaneously reconstruct the noisy poses and recognize human actions, and (2) Feature Aggregator (FA) to channel-wise aggregate the learned features: reconstructed poses and locations, actions, and egomotion to predict pedestrians’ future locations. }
		\label{fig:intro}
		%\vspace{-5mm}%Put here to reduce too much white space after your table 
	\end{figure}
	% \vspace{-5mm}%Put here to reduce too much white space after your table 
	
	Recent deep-learning-based methods\cite{mohamed2020social, wang2021graphtcn, chandra2020forecasting, alahi2016social} have shown promising prediction results in ‘perfect’ settings, where ground truth (or complete) observations are given. Using ground truth observations helps model human motion more accurately and may improve the prediction accuracies. However, the ground truth data is unavailable during test time. This limits the potential applicability of these methods in practice. Other methods~\cite{mangalam2020disentangling, yagi2018future} rely on pre-processing techniques to denoise the observations in advance of testing.  These approaches mainly focus on pre-processing (i.e. reconstructing or denoising) the human skeleton, an important feature for prediction. However, they are easily prone to errors under harsh conditions, such as fast camera motion and occlusions, especially in dynamic scenes. In this work, the following challenges are addressed: (1) reconstruction of human pose, which is a non-trivial task in computer vision. To the best of our knowledge, none of existing methods successfully reconstruct human skeletons in dynamic video sequences by exploiting the structural properties of human skeletons spatially and temporally. (2) the use of low-level human pose features to learn the higher-level action features. So far, the skeleton-based action features have not been considered for prediction tasks. 
	
	We design GPRAR to predict human future trajectory  under noisy observations in dynamic video scenes by devising solutions to the above challenges. It consists of two novel sub-networks: (1) a human pose reconstruction and action recognition network (PRAR) and (2) an encoder-decoder based Feature Aggregator (FA), shown in Figure~\ref{fig:intro}. The underlying idea of PRAR is to reconstruct human poses and learn action features simultaneously from the noisy pose detections. To best exploit the coherent and structural properties of human skeletons, PRAR is implemented with an encoder and two decoders, where each encoder and decoder is a multi-layer spatiotemporal graph convolutional network operating on the naturally connected human joints (or pose graph). Furthermore, we propose an encoder-decoder FA to channel-wise aggregate the learned features: reconstructed poses and locations, actions, and camera motion using temporal convolutional networks (TCNs). The aggregated feature is then used to output the future trajectory of a target pedestrian. In summary, the contributions of this paper are as follows: 
	\begin{itemize}
		\item We propose an efficient and robust human trajectory prediction network (GPRAR) under noisy observations (Section~\ref{sec:system_design}). GPRAR consists of two novel sub-networks: a human pose reconstruction and action recognition network (PRAR) and (2) an encoder-decoder based Feature Aggregator (FA).
		\item We evaluate our model on two commonly used datasets: TITAN and JAAD, and show that our method outperforms other methods with a large margin under noisy scenarios (Section~\ref{sec:experiments}). We also conduct ablation studies to demonstrate the effectiveness of each system component. 
	\end{itemize}
	%------------------------------------------------------------------------
	\section{Related Work}
	\label{sec:related_work}
	\noindent \textbf{Trajectory Prediction in Dynamic Scenes.} Most of the recent works in this research~\cite{malla2020titan,wang2021graphtcn,alahi2016social,yagi2018future,liang2019peeking,mangalam2020disentangling} rely on methodologies such as recurrent neural networks (RNNs)~\cite{mikolov2010recurrent}, temporal convolutional networks (TCNs)~\cite{lea2017temporal}, or their variants~\cite{hochreiter1997long, chung2014empirical}, which aggregate various input features during an observation period to model the relative human motion to the camera view. For instance, graph structures~\cite{wang2021graphtcn, mohamed2020social} can formulate interactions between agents (e.g. pedestrians, vehicles) using their past locations to predict the  future trajectory. Pedestrians’ shapes, scales, locations, and camera motion can be integrated using TCNs~\cite{yagi2018future}. Recently, human action has also been used for trajectory prediction task. For example, Malla et al.~\cite{malla2020titan} extracts action feature from the Two-Stream Inflated 3D ConvNet (I3D) and uses it as an input to their prediction model. Liang et al.~\cite{liang2019peeking} designs a complex two-branch network to simultaneously predict human future activities and trajectories. While the above methods have shown promising prediction results, they rely on ground truth human locations and pose features that are not available in real-world settings. 
	
	\noindent \textbf{Neural Networks on Graphs.} Spatial-temporal graph convolutional neural networks (ST-GCN) is originally proposed by Kipf et al.~\cite{kipf2016semi}, which extends the convolution operations from Convolutional Neural Network (CNN) to graph. ST-GCN and its variants have been widely used to model spatioltemporal features, which benefit several applications such as scene graph generation~\cite{gu2019scene}, point cloud classification and segmentation~\cite{landrieu2018large} , action recognition~\cite{carreira2017quo}, and semantic segmentation~\cite{lea2017temporal}. For the task of trajectory prediction, ST-GCN~\cite{wang2021graphtcn, mohamed2020social} is mainly used to model social interactions in static videos given the ground truth pedestrians’ locations. Our work, instead, tackles the prediction problem in dynamic video sequences, where the social interactions become less effective given the dynamic changes of a front-view camera. Another model which is technically related to ours is skeleton-based ST-GCN~\cite{yan2018spatial}.  However, this model is originally designed for action recognition task and assumes that complete human skeletons are available as inputs. To the best of our knowledge, GPRAR is the first prediction model that leverages graph convolutional network to simultaneously model robust human poses and action features for trajectory prediction under noisy observations.
	
	\section{System Design}
	\label{sec:system_design}
	\noindent \textbf{ System Overview.} As illustrated in Figure~\ref{fig2}, the task is to predict the future locations of $N$ pedestrians given the past $T_{obs}$ frames. For simplicity, let us describe this task for a target pedestrian $i \in N$ as follows:  
	\begin{figure*}[t]
		\centering
		%\fbox{\rule{0pt}{2in} \rule{.9\linewidth}{0pt}}
		\includegraphics[width=\textwidth]{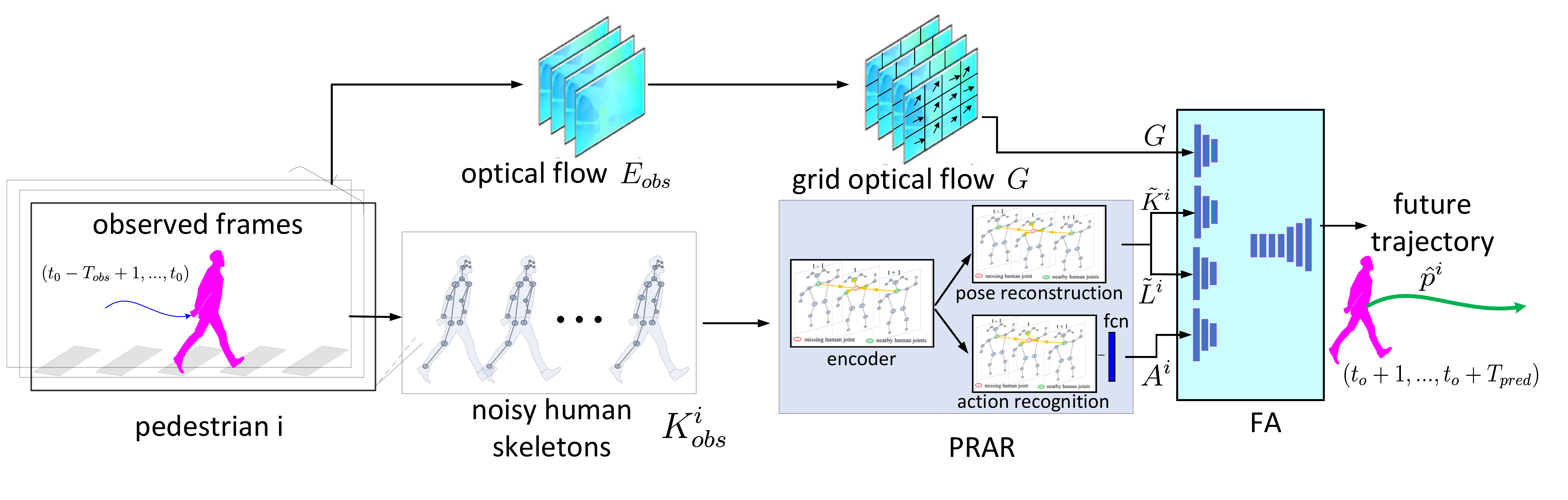}
		
		\caption{\textbf{GPRAR System Overview.} Our prediction model consists of two sub-networks: PRAR and FA.  Given the sequence of observed noisy human skeletons $K^i_{obs}$ of the target pedestrian $i$ as input, PRAR reconstructs (denoises) the noisy human skeletons and recognizes the human action. In the later stage, FA aggregates the learned features: grid optical flow $G$, reconstructed human skeletons $\tilde{K}^i$, locations $\tilde{L}^i$, and action feature $A^i$ and predicts the future trajectory of pedestrian i.}
		\label{fig2}
		\vspace{-1mm}%Put here to reduce too much white space after your table 
	\end{figure*}
	% \vspace{-5mm}%Put here to reduce too much white space after your table 
	
	\noindent \textbf{1.} At each time (current frame) $t_0$, our model receives noisy features extracted from the past $T_{obs}$ frames as inputs and produces the future trajectory $\hat{p}^i$ of pedestrian $i$ in the next $T_{pred}$  frames. We use two sequences of noisy input features: human skeletons $K_{obs}^i$  and optical flow $E_{obs}$ obtained using available public detectors (e.g., OpenPose~\cite{cao2019openpose} and FlowNet~\cite{ilg2017flownet}).  We note that recent works~\cite{malla2020titan, mohamed2020social, yagi2018future, wang2021graphtcn, alahi2016social} rely on ground truth input features, but in reality, these features are not available during testing. We, instead, only use noisy features as inputs.
	
	\noindent \textbf{2.} Given the sequence of observed noisy human skeletons $K_{obs}^i$ of pedestrian $i$, PRAR, a novel encoder-decoder based spatiotemporal pose graph convolutional network, reconstructs (denoises) these noisy human skeletons and recognizes the human actions. The sequence of reconstructed pose $\tilde{K}^i$ and action feature $A^i$ are then used as inputs to Feature Aggregator (FA) to predict the future locations. Moreover, since the observed location is an important feature to represent the overall human movements, we extract the sequence of reconstructed locations $\tilde{L}^i$  from the sequence of reconstructed pose $\tilde{K}^i$ and forward it to FA in the next stage. On the other branch (Figure~\ref{fig2}), we calculate the grid optical flow $G$, by dividing each optical flow image $e_t \in E_{obs}$ into grids of $3 \times 4$ and averaging the values of all pixels in each grid cell. The grid optical flow represents the camera motion in different regions of the scene, thus provides more accurate camera motion compared to the pixel-level optical flow. In the last stage, Feature Aggregator (FA) aggregates all learned features $G,\tilde{K}^i,\tilde{L}^i,A^i$ via an encoder-decoder based one-dimensional temporal convolutional network to predict the future trajectory  $\hat{p}^i$ of pedestrian $i$  in the next $T_{pred}$ frames.  
	
	\textbf{Notations.} We denote the sequence of observed skeletons of pedestrian $i$  as $K_{obs}^i=\{\mathbf{k}_t^i, \forall t=\{t_0 - T_{obs} + 1,…,t_0\}\}$,  where  $\mathbf{k}_t^i=\{v_{kt}^i,\forall k=\{1…,\mathcal{K}\}\}$ is a human skeleton consisting of $\mathcal{K}$ human joints $v_{kt}^i$ of pedestrian $i$ at time $t$.  The initial value of human joint $v_{kt}^i$ is $f(v_{kt}^i )= (x_{kt}^i,y_{kt}^i,c_{kt}^i)$, where $(x_{kt}^i,y_{kt}^i)$ is the two-dimensional coordinate and $c_{kt}^i$  is the confidence value of that joint. We note that not all human joints are visible due to occlusions or poor performance of the human detector. Thus, we set the initial value of missing human joints to $(0,0,0)$. The sequence of reconstructed human skeletons $\tilde{K}^i$ has the same size as $K_{obs}^i$. The sequence of observed optical flow is denoted as $E_{obs}=\{e_t, \forall t \in \{t_0-T_{obs}+1 ,…,t_0\}\},$ where $e_t$ is the optical flow image with dimension $W \times H$ (width x height).  We note that while the sequence of human skeletons $K_{obs}^i$ is unique for each pedestrian, the flow $E_{obs}$  is shared for all pedestrians during the same observation time. The sequence of reconstructed locations is written as $\tilde{L}^i=\{(\tilde{x}_t^i,\tilde{y}_t^i ), \forall t=\{t_0-T_{obs}+1,…,t_0\}\}$, where $(\tilde{x}_t^i,\tilde{y}_t^i )$ is the two-dimensional coordinate of the middle hip of human skeleton of pedestrian $i$ at time $t$. The learned action feature of pedestrian $i$ is denoted as $A^i=\{a_t^i,\forall t=\{t_0-T_{obs}+1,…,t_0\}\}$, where $a_t^i$  is the action feature in each frame $t$. Lastly, the predicted trajectory of pedestrian i is denoted as $\hat{p}^i=\{(\hat{x}_t^i,\hat{y}_t^i ), \forall t=\{t_0+1,…,t_0+T_{pred}\}$, where $(\hat{x}_t^i,\hat{y}_t^i)$ is the predicted location in  the future frame.  Next, we present the intuitions and design details of our prediction networks: PRAR (Section~\ref{subsec:pose_rec}), FA (Section~\ref{subsec:fa}). 
	
	\begin{figure*}
		\begin{subfigure}[t]{0.55\textwidth}
			\includegraphics[width=85mm, height=66mm]{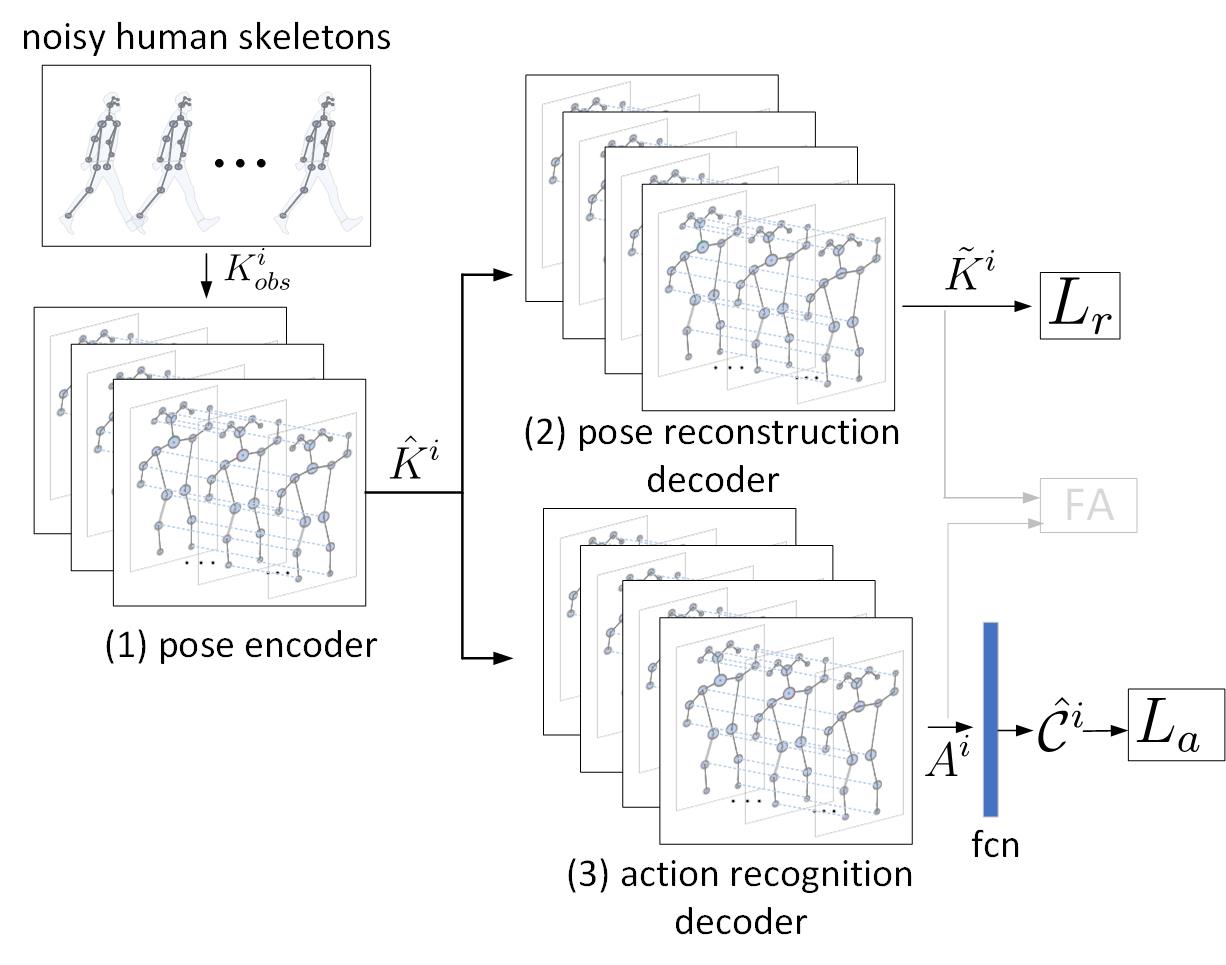}
			\caption{PRAR model overview}
			\label{fig:PRARa}
		\end{subfigure}
		\begin{subfigure}[t]{0.45\textwidth}
			\includegraphics[width=70mm, height=66mm]{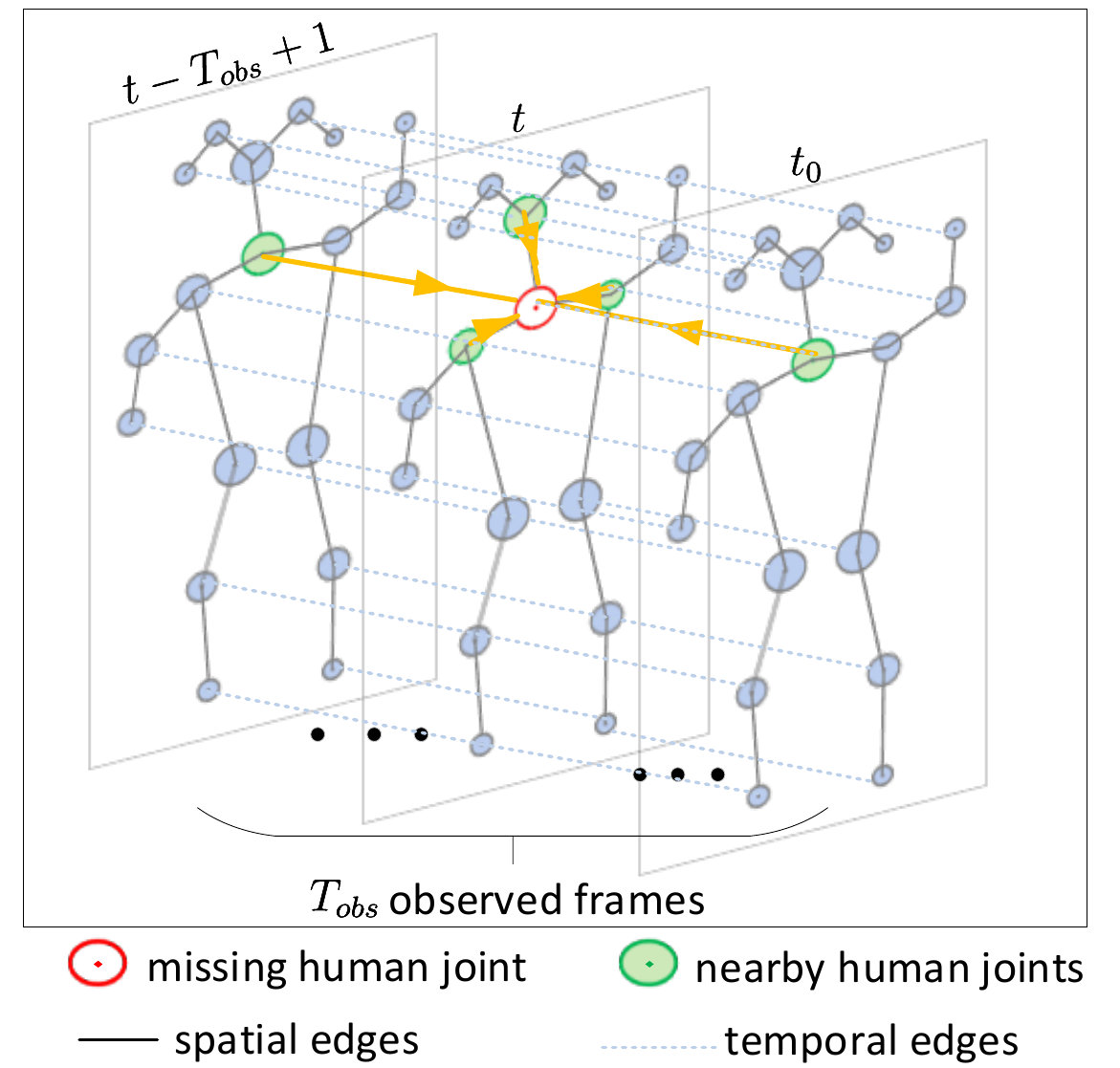}
			\caption{PRAR's single layer}
			\label{fig:PRARb}
		\end{subfigure}
		\caption{\textbf{PRAR Network Architecture.} (a) PRAR consists of a pose encoder and two decoder branches for pose reconstruction and action recognition. (b) A single layer of each pose encoder/decoder is designed  using the skeleton-based spatial-temporal graph convolutional network. An example of reconstructing the missing joint is also shown in (b), where it can be achieved by considering the information from nearby joints in both spatial and temporal domains.}
		\label{fig:PRAR}
	\end{figure*}
	
	\subsection{Pose Reconstruction and Action Recognition Network (PRAR)}
	\label{subsec:pose_rec}
	The goal of PRAR is to generate robust action and pose features that boost the prediction under extreme scenarios (e.g., occlusions, fast camera motion, etc.) and those that compromise the susceptibility of a public detector.  To this end, we design PRAR consisting of an encoder and two decoders, each of which comprises multi-layer of the graph neural networks (GNNs) operating on the sequence of observed human skeletons, shown in Figure~\ref{fig:PRAR}. Before discussing the details, let us highlight PRAR with the following technical novelties: (1) while some works~\cite{cheng2020skeleton, yan2018spatial} use GNNs on human skeletons solely for the task of action recognition, GNNs have not been utilized/extended for the pose reconstruction and human trajectory prediction tasks. (2) To the best of our knowledge, PRAR is the first encoder-decoder based GNNs for multi-task learning, in which robust pose features are learned to benefit both tasks mentioned above.  (3) PRAR is a plug-and-play module that is trained separately but also can be integrated and jointly trained with other models for human trajectory prediction. We illustrate this training setup in Section~\ref{sec:experiments}.
	
	As shown in Figure~\ref{fig:PRARa}, PRAR consists of three main components. (1) A pose encoder that takes the sequence of noisy observed human skeletons $K_{obs}^i$ as input and learns comprehensive encoded pose features $\hat{K}^i$. (2) A pose reconstruction decoder that takes the encoded pose features $\hat{K}^i$ and produces a complete (denoised) sequence of human skeletons $\tilde{K}^i$. (3) An action recognition decoder that also uses the shared pose feature $\hat{K}^i$ to generate action feature $A^i$, which is then input to a fully connected network (fcn) to generate an action class $\hat{\mathcal{C}}^i$ of pedestrian $i$. While the learned features $\tilde{K}^i$ and $A^i$ are forwarded to FA module for trajectory prediction, $\tilde{K}^i$ and $\hat{\mathcal{C}}^i$ are used to train PRAR separately at the pre-training phase using reconstruction loss $L_r$  and action recognition loss $L_a$.  
	
	Since each encoder/decoder consists of multiple layers of GNNs, let us first present the details of a single layer $(l)$, illustrated in Figure~\ref{fig:PRARb}. In each layer, we model the sequence of observed human poses as a spatial-temporal skeleton-based graph $G =(V,E)$, where the nodes $V=\{v_{kt}, \forall t \in \{t_0-T_{obs}+1,…,t_0\}, \forall k \in \{1,…,\mathcal{K}\}\} $  are all the human joints in a skeleton sequence. The edge set $E$ consists of a spatial edge set $E_{sp}=\{v_{kt}v_{jt},(k,j) \in \mathcal{K}\}$ that connects human joints naturally within a frame, and a temporal edge $E_{tp}=\{v_{kt}v_{k(t+1)}, \forall t \in \{t_0-T_{obs}+1,…,t_0\}, \forall k \in \{1,…,\mathcal{K}\}\}$ that connects the same human joint in consecutive observed frames. We note that Yan et al.~\cite{yan2018spatial} and Cheng et al.~\cite{cheng2020skeleton} utilize similar skeleton-based graph neural networks. However, these works do not apply to our case because they only focus on modeling human actions. PRAR is designed not only to learn a spatiotemporal action feature, but also to reconstruct the human missing joints. The reconstruction task is done through spatial-temporal graph convolutions by leveraging the message passing mechanism, in which directly-connected nodes can communicate with each other. Specifically, the spatial-temporal graph convolution is utilized to estimate the coordinate of a missing human joint using the information of nearby human joints in both spatial and temporal domains. In Figure~\ref{fig:PRARb}, we assume human neck as a missing human joint to illustrate this reconstruction process in details. The coordinate of a missing human joint at the observed frame  $t \in \{t_0-T_{obs}+1,…,t_0\}$ is calculated by gathering information from nearby visible human joints (nose, right shoulder, and left shoulder) within the same frame $t$  and information of the same joint from neighboring observed frames. In general, given the input feature of a missing node of pedestrian $i$ at layer $(l)$ is $f(v_{kt}^{i(l)})$, we apply the spatial-temporal graph convolution operation to estimate the new value  $\tilde{f}(v_{kt}^{i(l)})$ of the node $v_{kt}^{i(l)}$ as:
	\begin{equation}
		\tilde{f}(v_{kt}^{i(l)}) = \sum_{v^{i(l)}_{jt} \in B(v_{kt}^{i(l)})} \frac{1}{z_t^{i(l)}}f(v_{jt}^{i(l)})w_{jt}^{i(l)}  
		\label{Eq:conv}
	\end{equation}
	
	\noindent where $B(v_{kt}^{i(l)})$ is the set of nearby human joints $v_{jt}^{i(l)}$ that naturally connect to the node $v_{kt}^{i(l)}$ in both spatial and temporal axes; $w_{jt}^{i(l)}$ is a learnable weight vector; $z_t^{i(l)}$ is a normalization term, which normalizes the output features to range $[0, 1]$. We note that the initial value of the missing node $v^i_{kt}$ at layer $(0)$ is $ f(v_{kt}^{i(0)}) = (0,0,0)$, which is the noisy output of a human pose detector. The output feature $\tilde{f}(v_{kt}^{i(l)})$ at layer $(l)$ is forwarded to the next layer $(l+1)$  to calculate $\tilde{f}(v_{kt}^{i(l+1)})$. In the last layer of the pose reconstruction decoder, $\tilde{f}(v_{kt}^{i}) = (\tilde{x}_{kt}^{i},\tilde{y}_{kt}^{i},\tilde{c}_{kt}^i)$ where $(\tilde{x}_{kt}^{i},\tilde{y}_{kt}^i)$ is the two-dimensional reconstructed location and $\tilde{c}_{kt}^i$ is the new confidence score. Note that Equation~\ref{Eq:conv} can be applied for other non-missing nodes as well. Interestingly, we found that using the spatiotemporal graph convolution for all nodes improves the prediction results as it enhances the coherency of human skeletons.
	
	\textbf{Encoder and Decoders.} Although the spatial-temporal graph convolution in a single layer is useful for reconstructing human poses, it only considers the information from the nearby nodes, which directly connect to the missing node. In fact, the other non-directly connected nodes may also have impacts on the missing nodes (e.g., given pedestrian’s head locations, we, as humans, can estimate the locations of non-directly connected legs). Based on this intuition, we design each encoder and decoder with multiple layers of spatial-temporal graphs as multi-layers allow non-directly connected nodes to have impacts on the target node. Specifically, we use three layers for encoder and four layers for each decoder. These specific numbers of layers are determined based on our empirical study, in which we achieved saturated prediction results. We discuss the details of these network parameters in the supplementary materials. 
	
	As a result, the last layer of pose reconstruction decoder produces the reconstructed pose $\tilde{K}^i$. The last layer of action recognition decoder outputs the learned action feature $A^i$, which is then forwarded to fcn to calculate the action label $\hat{\mathcal{C}}^i$. While the action label $\hat{\mathcal{C}}^i$ is used to optimize PRAR for action recognition task, $A^i $ is used as input to FA in the next stage.

	\textbf{Losses. }PRAR is initially trained separately from our prediction model (at pre-training phase) with the proposed multi-task loss as below: 
	\begin{equation}
		L=L_r+L_a
		\label{Eq:L}
	\end{equation}
	where $L_r$ is pose reconstruction loss and $L_a$ is action recognition loss. For pose reconstruction loss, we use mean-square-error loss over predicted human joints as follows:
	\begin{equation}
		L_r=  \sum_{i}^{N}\sum_{k}^{K}\sum_{t}^{T_{obs}} \norm{\tilde{f}(v_{kt}^i)-\bar{f}(v_{kt}^i)}^2
		\label{Eq:Lr}
	\end{equation}
	where $\tilde{f}(v_{kt}^i)$ and $\bar{f}(v_{kt}^i)$ are reconstructed and ground truth values of human joint $v_{kt}^i$ . For action recognition loss, we use the cross-entropy loss: 
	\begin{equation}
		L_a= \sum_i^Nce(\hat{\mathcal{C}}^i, \bar{\mathcal{C}}^i)
		\label{Eq:La}
	\end{equation}
	where $ce$ is cross-entropy function; $\hat{\mathcal{C}}^i$ and $ \bar{\mathcal{C}}^i$ are the predicted and ground truth action class labels of pedestrian $i$, respectively.
	
	\subsection{Feature Aggregator (FA)}
	\label{subsec:fa}
	The goal of FA is to channel-wise aggregate all the learned features: reconstructed pose $\tilde{K}^i$, reconstructed location $\tilde{L}^i$, action feature $A^i$, and regional optical flow $G$. To this end, we design an encoder-decoder based on temporal neural networks, where the encoder channel-wise aggregates these features, the decoder takes the aggregated (encoded) feature as input and generates the future trajectory $\hat{p}^i$. Although Yagi et al.~\cite{yagi2018future} (FPL) adopted a closely related network architecture, our FA is a much smaller network that outperforms FPL. Specifically, the number of network parameters of FA is 33\% less than FPL’s even though FA accommodates more input features. We believe this is due to the effectiveness of PRAR, which has produced the robust learned features.  The model details of FA are provided in the supplementary materials. 
	
	Once PRAR is trained separately for pose reconstruction and action recognition using loss function in Equation~\ref{Eq:L}, the entire prediction network is trained using mean-square-error loss as follows: 
	\begin{equation}
		L(W)= \sum_i^N \sum_t^{T_{pred}} \norm{\hat{p}_t^i - \bar{p}_t^i}^2
		\label{Eq:pred}
	\end{equation}
	in which $W$ includes all the trainable parameters of the model, $\hat{p}_t^i$ and $ \bar{p}_t^i$ are the predicted and ground truth locations of pedestrian $i$ at time $t$, respectively.
	
	%------------------------------------------------------------------------
	
	\section{Experiments}
	\label{sec:experiments}
	\noindent \textbf{Datasets.} We evaluate our model on two publicly available autonomous driving datasets: JAAD~\cite{kotseruba2016joint}   and TITAN~\cite{malla2020titan}. JAAD contains 346 videos with 82,032 frames. The videos are recorded by a front-view wide-angle camera mounted in the center of the front windshield. These videos range from short 5-second clips to 15-second videos shot under various scenes, weathers, and lighting conditions. TITAN consists of 700 video clips with 75,262 annotated frames and 395,770 persons. The annotations provide a variety of pedestrians’ actions and interactions.  We use 9 action classes (standing, jumping, squatting, bending, running, walking, laying down, sitting, kneeling) provided by TITAN, while JAAD supports two action classes: walking and standing. We conduct experiments in each dataset separately with the train/validation ratio of 80/20.
	
	\noindent \textbf{Training Setup.}  Training is done in two stages: (1) PRAR plays a vital role in our prediction network; to be impactful, it is crucial to successfully train PRAR for pose reconstruction and action recognition before (2) customizing it for the prediction task. The details are as follows.
	
	\noindent  \textbf{Stage 1:} As TITAN and JAAD have limited numbers of complete human skeletons, we first train PRAR on Kinetics dataset~\cite{kay2017kinetics} to obtain the initial network weights. Kinetics is a large human skeleton dataset, which contains around 300,000 video clips with 400 human action classes of various daily activities. Once PRAR is successfully trained on Kinetics dataset, we continue training the pre-trained PRAR on TITAN and JAAD datasets.  The trained PRAR model on TITAN and JAAD datasets (obtained from the last training epoch) is used in the next training stage.
	
	\noindent  \textbf{Stage 2:} We customize the pre-trained PRAR to the trajectory prediction task on JAAD~\cite{kotseruba2016joint} and TITAN~\cite{malla2020titan} datasets. Specifically, we attach FA module on top of PRAR and train the entire prediction model using the loss function in Equation~\ref{Eq:pred}.  Our training setup is considered an adaptive learning approach as opposed to the non-adaptive learning where the input features to the predictor (i.e., FA in our work) are fixed. We show the effectiveness of this learning approach in the ablation study. 
	
	In both stages, our model is trained using stochastic gradient descent~\cite{bottou2012stochastic} with a learning rate of 0.01 and 50 epochs. We decay the learning rate by 0.1 after every 10 epochs. To implement spatial-temporal graph convolutions, we use similar implementation steps discussed in~\cite{kipf2016semi}. Our network model is implemented using PyTorch~\cite{paszke2019pytorch}. 
	
	\noindent  \textbf{Comparison Methods.} We compare our prediction results with two baselines (Const-vel~\cite{malla2020titan}, LSTM~\cite{alahi2016social}) and three state-of-the-art methods (Social-STGCNN~\cite{mohamed2020social}, FPL~\cite{yagi2018future}, and TITAN~\cite{malla2020titan}). (1) Const-Vel uses the velocity, calculated using the last two observed locations, to interpolate future positions linearly. (2) LSTM models individual pedestrian behaviors using LSTM cells. (3) Social-STGCNN models social interactions between pedestrians in the scenes by using graph neural networks. (4) FPL incorporates pose, scales, egomotion features using temporal neural networks. (5) TITAN uses action features in combination data from IMU sensors for predicting future locations.
	
	\noindent \textbf{Evaluation Metrics.} We evaluate our system using two commonly used metrics~\cite{yagi2018future, alahi2016social}: (a) average displacement error (ADE): mean square error over all locations of predicted and true trajectories; (b) final displacement error (FDE): mean square error at the final predicted and true locations of all human trajectories.
	
	Next, we present our pose reconstruction and action recognition results (Section~\ref{subsec:PRAR}), prediction results with ablation studies (Section~\ref{subsec:quanti} and~\ref{subsec:quali}).
	\begin{figure*}
		\begin{minipage}{0.45\textwidth}
			\centering
			\includegraphics[width=\textwidth]{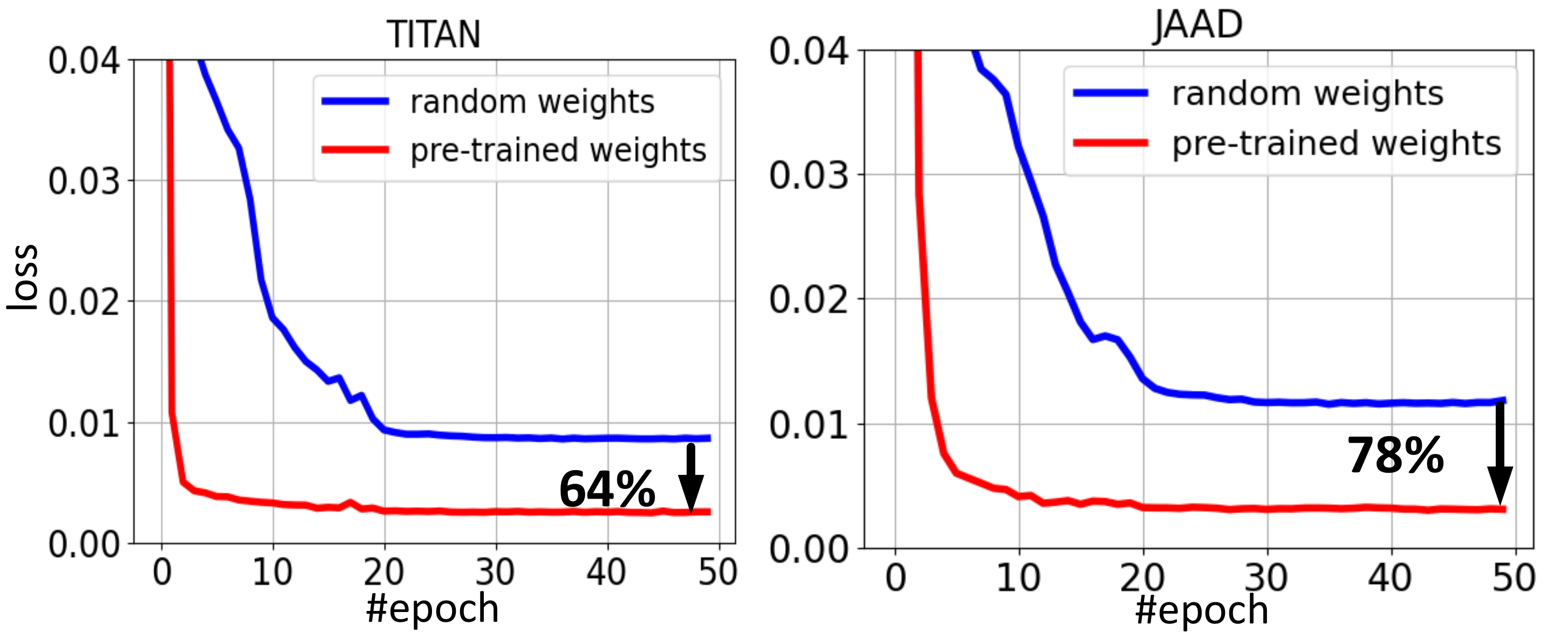}
			\caption{Effect of pre-training PRAR on Kinetics dataset. With the pre-trained network weights, pose reconstruction losses are decreased (better) 64\% and 78\% compared with using random weights on both datasets.}
			\label{fig:PRAR_plot}
		\end{minipage}
		\hspace{0.10mm}
		\begin{minipage}{0.49\textwidth}
			\centering
			\includegraphics[width=\textwidth]{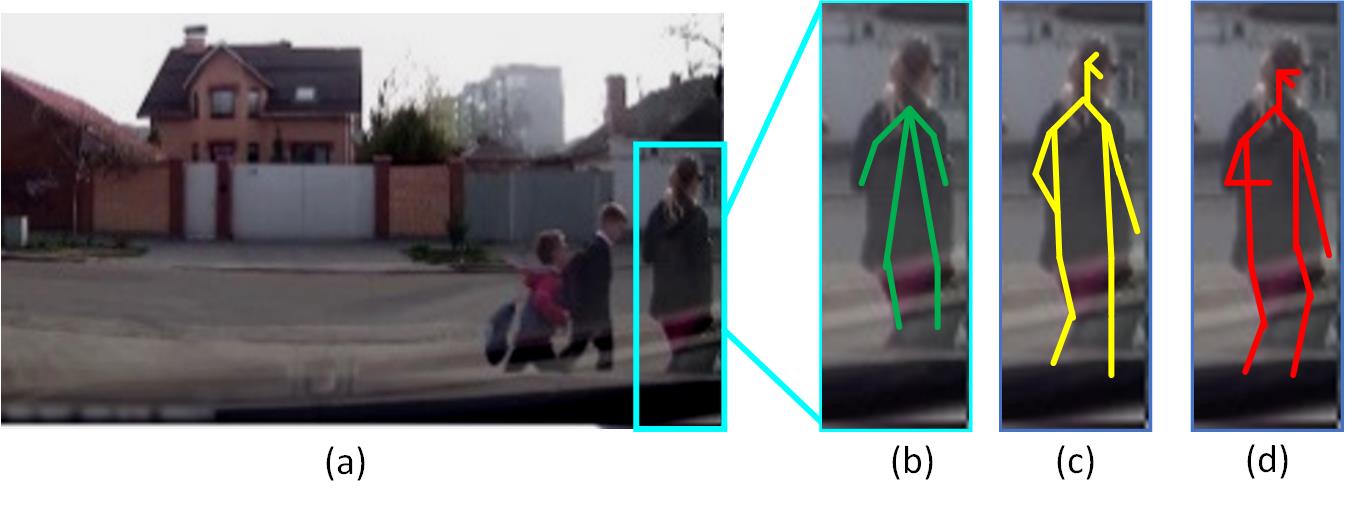}
			\caption{A pose reconstruction result on JAAD. (a) original frame with a target pedestrian inside the bounding box, (b) noisy pose detection with missing head and legs, (c) our reconstruction result, and (d) ground truth pose.}
			\label{fig:PRAR_quali}
		\end{minipage}
	\end{figure*}
	\begin{table*}[!ht]
		\centering
		\begin{tabular}{c |ccc| ccc}
			\toprule
			\multirow{2}{*}{Models}        & \multicolumn{3}{c|}{JAAD} 		  & \multicolumn{3}{c}{TITAN}  \\
			\cmidrule{2-4}
			\cmidrule{5-7}
			& N. 		       & Pr. 		 & C. 		   & N. 		   & Pr. 		 & C.\\
			\midrule
			Const-vel~\cite{malla2020titan}  	& 48.07/75.03      & 38.56/59.98 & 18.68/30.15 & 150.31/238.70 & 84.18/35.47 & 23.81/42.72 \\
			LSTM~\cite{alahi2016social}  			& 46.96/57.28      & 45.70/56.12 & 51.04/61.42 & 51.52/80.59   & 50.99/79.90 & 28.92/50.51 \\
			Social-STGCNN~\cite{mohamed2020social}    & 80.46/71.38      & 80.46/71.38 & 79.14/58.73 & 72.32/52.86   & 73.62/54.84 & 68.78/51.68 \\
			FPL~\cite{yagi2018future}    			& 27.07/26.92      & 28.01/31.22 & 24.85/27.31 & 52.26/80.93   & 34.13/50.92 & 14.09/19.98 \\ 
			TITAN~\cite{malla2020titan}* 			& - 				& -			& -			   & -       		& - & \textbf{11.32/19.53} \\
			GPRAR (Ours)     & \textbf{21.09/21.62}      & \textbf{18.13/20.88} & \textbf{14.79/20.38} & \textbf{26.17/38.58} & \textbf{24.49/34.85} & 12.56/20.36 \\
			\bottomrule
		\end{tabular}
		\caption{Quantitative results (ADE/FDE in pixels) on JAAD and TITAN datasets in different observation modes: noisy (N.), pre-processed (Pr.), and complete (C.). The lower the better. (*) Since the implementation is not publicly available, we use the results reported in~\cite{malla2020titan}.}
		\label{table:quanti}
	\end{table*}
	
	\begin{table}[!ht]
		\centering
		\begin{tabular}{lcc}
			\toprule
			Models          & ADE 		  & FDE 	\\
			\midrule
			X   			& 30.55      & 45.42 \\
			XR  			& 28.94      & 44.03 \\
			XR + C     		& 27.92      & 41.58 \\
			XR + A 			& 28.08      & 41.23 \\
			XR + PR     	& 28.13      & 41.94\\
			\midrule
			\textbf{Full Model} (XR + C + A +PR)   & 26.17      & 38.58 \\
			\bottomrule
		\end{tabular}
		\caption{Ablation Study. Effects of each feature used in our method (GPRAR): noisy location (X), reconstructed location (XR), camera motion (C), action (A), and reconstructed pose (PR).}
		\label{table:ablation}
	\end{table}	
	
	\subsection{Pose Reconstruction and Action Recognition Results. }
	\label{subsec:PRAR} 
	We found that pre-training PRAR on Kinetics dataset (in stage 1) significantly improves the pose reconstruction losses for both TITAN and JAAD datasets as depicted in Figure~\ref{fig:PRAR_plot}.  Interestingly, by using the pre-trained network weights, pose reconstruction losses are significantly decreased  (better) by 64\% and 78\% on TITAN and JAAD respectively in comparison with using random weights.  
	
	Figure~\ref{fig:PRAR_quali} shows an example qualitative result of PRAR on JAAD validation data. We observe that PRAR is capable of reconstructing the missing human joints (e.g., missing head and legs in this figure). Quantitatively, we achieve pose reconstruction error of about $5$ pixels and $10$ pixels on TITAN and JAAD datasets given the image dimensions $1080 \times 1080$ pixels. 
	
	For the action recognition task, PRAR achieved 99\% accuracy on JAAD with two action classes, and 91.05\% on TITAN with eight action classes. We consider these to be desirable accuracies for skeleton-based action recognition. 
	
	\subsection{Quantitative Prediction Results on JAAD and TITAN Datasets. }
	\label{subsec:quanti}
	Table~\ref{table:quanti} shows quantitative results on JAAD and TITAN datasets (results from training stage 2). We compare our method (GPRAR) with other methods in three different observation modes: noisy, pre-processed, and ground truth. In the noisy (raw) mode, the observed data (poses and locations) are the outputs of a pose detector.  In the pre-processed mode, the data are estimated using KNN-imputer as used in~\cite{yagi2018future}. The ground truth observation is the complete pose data with no missing joints. Our model outperforms other methods on JAAD dataset in all three different scenarios. Specifically, our prediction results are 50\% and 22\% better than FPL in the noisy mode on TITAN and JAAD datasets, respectively. In the ground truth mode, our model outperforms others on JAAD dataset and produces very close results to TITAN. However, we note that TITAN method uses the IMU sensor data as additional features for prediction, while our method only relies on image data.
	
	\noindent \textbf{Ablation study.} The importance of individual features in GPRAR model is shown in Table~\ref{table:ablation}. We observe that using reconstructed location (XR) improves the prediction accuracy in comparison with using noisy location (X). Moreover, using reconstructed location in combination with other features: action (A), camera motion (C) and reconstructed pose (PR) further improves the results. Our full model, which aggregates all features, achieves the best prediction results. 
	
\begin{figure}[!t]
		
		\centering
		\includegraphics[width=65mm, height=45mm]{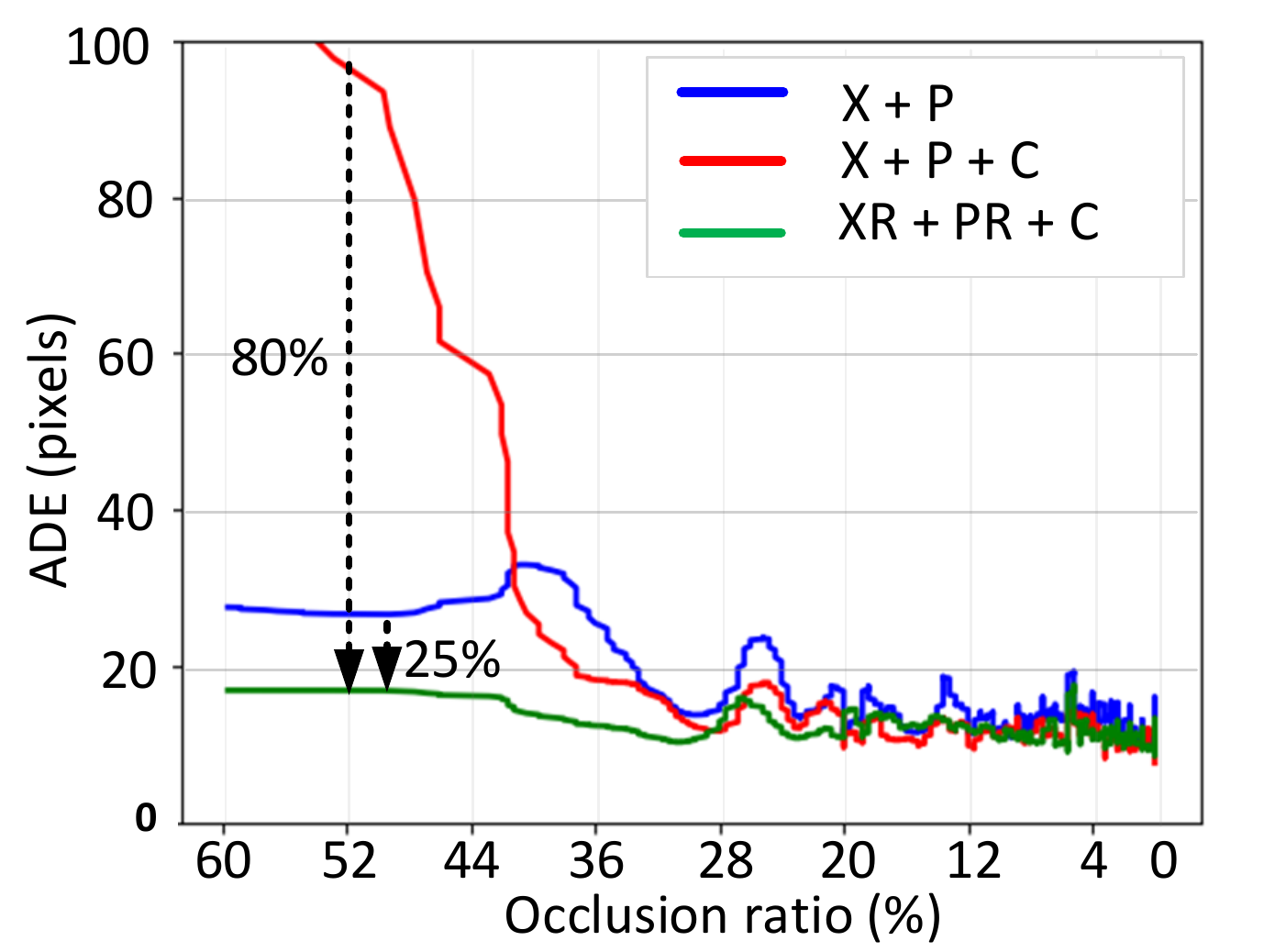}
		\caption{Impact of human occlusions on the prediction error.}
		\label{fig:impact_occlusion}
	\end{figure}

	\noindent \textbf{Impact of occlusions.} To observe the impact of occlusions on performance of GPRAR, we compare three different combination of features: (a) noisy location and noisy pose (X + P); (b) noisy location, noisy pose, and camera motion (X + P + C); (c) reconstructed location, reconstructed pose, and camera motion (XR + PR + C). As shown in Figure~\ref{fig:impact_occlusion}, all three variants perform well when the occlusion ratio is low. However, the performances of both (X + P) and (X + P + C) are significantly degraded under high occlusion scenarios, while the the proposed method with reconstructed pose (XR + PR + C) is less impacted by the occlusions and shows a steady good result (low ADE). Specifically, (XR + PR + C) reduces the prediction error (ADE) by 80\% and 25\% compared to  (X + P + C) and (X + P) at the occlusion ratio of 50\% (i.e. a half human body is occluded).
	\begin{figure*}[!t]
		\centering
		%\fbox{\rule{0pt}{2in} \rule{.9\linewidth}{0pt}}
		\includegraphics[width=\textwidth]{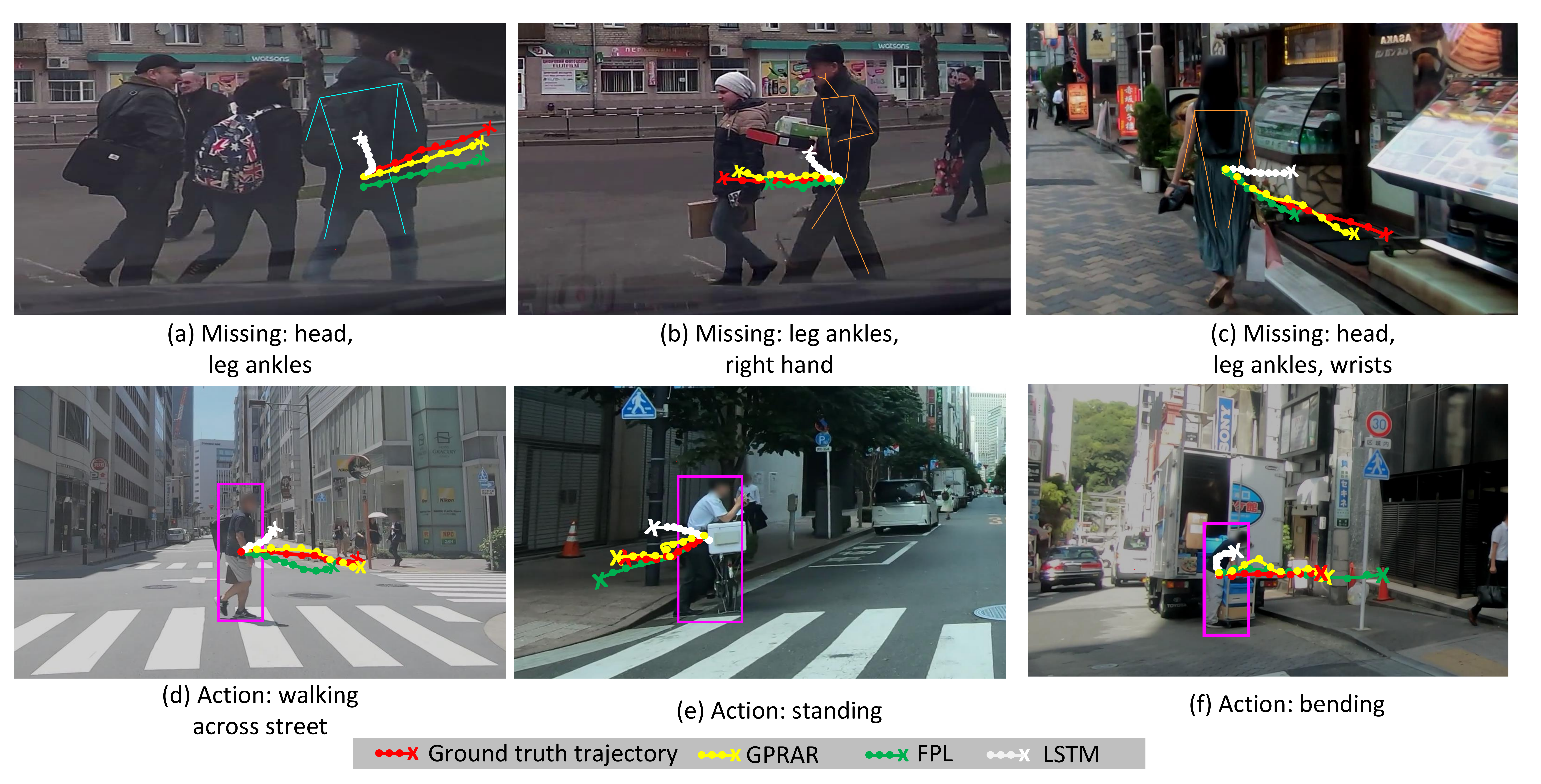}	
		
		\caption{Qualitative results of our model (GPRAR) in comparisons with other models: FPL, LSTM for scenarios of missing human joints (top row) and action types (bottom row). Additional results are provided in the supplementary materials.}
		\label{fig:pred_qual}
		\vspace{-3mm}%Put here to reduce too much white space after your table 
	\end{figure*}
	
\begin{figure}[!th]
		\centering
		\includegraphics[width=65mm, height=45mm]{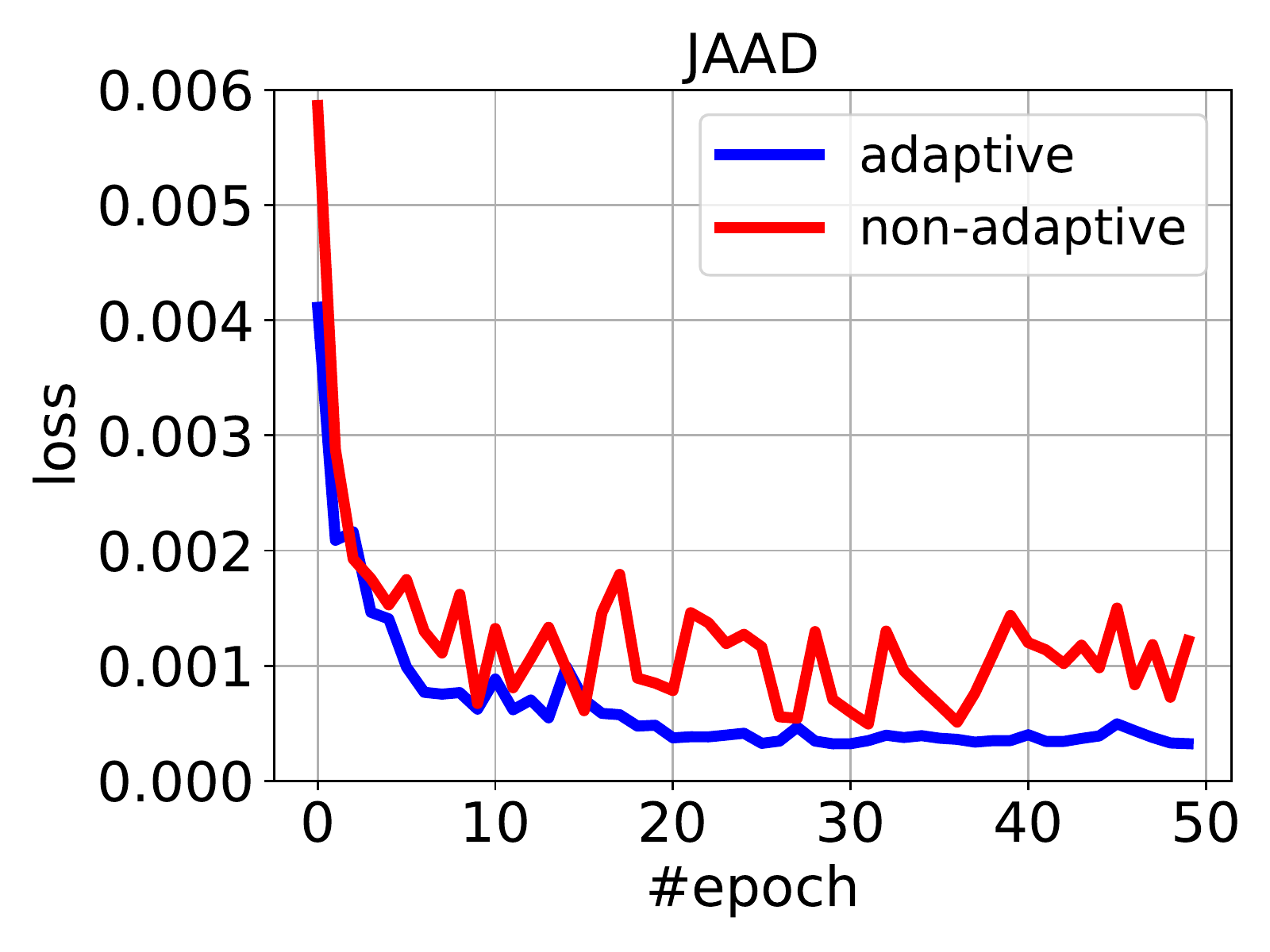}
		\caption{The effectiveness of our adaptive learning approach.}
		\label{fig:adaptive_learning}
		\vspace{-3mm}%Put here to reduce too much white space after your table 
	\end{figure}

	\noindent \textbf{Impact of adaptive learning.} Empirically, we found that the adaptive learning approach effectively improves the prediction accuracy.  Figure~\ref{fig:adaptive_learning} shows the effectiveness of this adaptive approach during training. It produces a lower stable loss in comparison with the non-adaptive one.
	\subsection{Qualitative Results.}
	\label{subsec:quali}
	Figure~\ref{fig:pred_qual} shows sample qualitative results comparing GPRAR with FPL and LSTM in various scenarios of noisy poses (Figure~\ref{fig:pred_qual}, top row) and action types (Figure~\ref{fig:pred_qual}, bottom row). Due to the pose reconstruction capability, GPRAR outperforms others under occlusion scenarios, such as: (a) the pedestrian is too close to the car; thus, there are missing head and ankles; (b) the pedestrian walks across the street with hidden right hand (c) the pedestrian walks away with missing face and ankles (not detected). As GPRAR also considers the skeleton-based action feature, GPRAR outperforms others in different human action scenarios: (d) walking across street, (e) standing and waiting, and (f) bending while unloading packages.
	
	\section{Conclusions}
	In this paper, we present GPRAR, a novel human future trajectory prediction model in dynamic video sequences, which efficiently handles noisy real-world scenarios. The main contributions of this paper are two novel subnetworks: PRAR and FA. While PRAR is trained for multi-task learning: action recognition and pose recognition, FA aggregates multiple learned features for trajectory prediction. The key implementation of PRAR is using the encoder-decoder based graph convolutional neural networks, which help exploit the structural properties of human poses. Through extensive experiments, we have shown GPRAR produces superior performance in comparisons with the state-of-the-art models. We have further presented performance analysis of introduced features and shown that our model performs effectively under occlusions and various human actions.
	
	{\small
		\bibliographystyle{ieee_fullname}
		\bibliography{egbib}
	}
	
\end{document}